\documentclass[final,12pt]{clear2025} %

\usepackage[utf8]{inputenc}
\usepackage[T1]{fontenc}
\usepackage{graphicx}
\usepackage{xcolor}
\usepackage{microtype}
\usepackage[parfill]{parskip}
\usepackage{url}            %
\usepackage{amsmath,amssymb,amsfonts}
\usepackage{bm}
\usepackage{xpatch}
\usepackage{tikz}
\usepackage{algorithm}
\usepackage[noend]{algpseudocode}
\usepackage{mathtools}
\usepackage{kbordermatrix}
\usepackage[capitalize,noabbrev]{cleveref} 
\usepackage{multirow}  
\usepackage{booktabs}
\usepackage{enumitem}
\usepackage{fancyhdr}

\graphicspath{{figs/}}
\usetikzlibrary{cd,arrows,calc,shapes,positioning}
\tikzstyle{obs} = [circle,fill=white,draw=black,inner sep=0pt,minimum size=18pt,font=\fontsize{10}{10}\selectfont,node distance=1,thick]
\tikzstyle{latent} = [obs,dotted]

\renewcommand{\kbldelim}{(}%
\renewcommand{\kbrdelim}{)}%

\newcommand{\xhdr}[1]{\textbf{#1}\:}

\newcommand{\bR}{\ensuremath \mathbb{R}}
\newcommand{\bN}{\ensuremath \mathbb{N}}

\newcommand{\cG}{\ensuremath \mathcal{G}}

\newcommand{\cE}{\ensuremath \mathcal{E}}
\newcommand{\cF}{\ensuremath \mathcal{F}}

\newcommand{\cL}{\ensuremath \mathcal{L}}

\newcommand{\cD}{\ensuremath \mathcal{D}}
\newcommand{\cQ}{\ensuremath \mathcal{Q}}

\newcommand{\bV}{\ensuremath \bm{V}}
\newcommand{\bE}{\ensuremath \bm{E}}
\newcommand{\paG}{\ensuremath \mathrm{pa}_{\cG}}
\newcommand{\pa}{\ensuremath \mathrm{pa}}
\newcommand{\N}{\ensuremath \bm{N}}

\newcommand{\lb}{l}
\newcommand{\ub}{u}

\DeclareMathOperator{\tr}{\ensuremath \mathrm{tr}}

\newcommand{\given}{\,|\,}
\newcommand{\E}{\mathbb{E}}

\newcommand{\X}{\ensuremath \bm{X}}
\newcommand{\x}{\ensuremath \bm{x}}
\newcommand{\Y}{\ensuremath \bm{Y}}
\newcommand{\y}{\ensuremath \bm{y}}
\newcommand{\Z}{\ensuremath \bm{Z}}
\newcommand{\z}{\ensuremath \bm{z}}

\Crefname{equation}{Eq.}{Eqs.}
\Crefname{figure}{Fig.}{Figs.}
\Crefname{tabular}{Tab.}{Tabs.}
\Crefname{theorem}{Thm.}{Thms.}
\Crefname{lemma}{Lem.}{Lems.}
\Crefname{proposition}{Prop.}{Props.}
\Crefname{definition}{Def.}{Defs.}
\Crefname{algorithm}{Alg.}{Algs.}
\Crefname{corollary}{Corol.}{Corol.}
\Crefname{section}{Sec.}{Sec.}

\makeatletter
\newcommand{\xdashleftrightarrow}[2][]{\ext@arrow 3359\leftrightarrowfill@@{#1}{#2}}
\def\rightarrowfill@@{\arrowfill@@\relax\relbar\rightarrow}
\def\leftarrowfill@@{\arrowfill@@\leftarrow\relbar\relax}
\def\leftrightarrowfill@@{\arrowfill@@\leftarrow\relbar\rightarrow}
\def\arrowfill@@#1#2#3#4{%
  $\m@th\thickmuskip0mu\medmuskip\thickmuskip\thinmuskip\thickmuskip
   \relax#4#1
   \xleaders\hbox{$#4#2$}\hfill
   #3$%
}
\makeatother

\makeatletter
\xpatchcmd{\algorithmic}{\itemsep\z@}{\itemsep=0.5ex plus1pt}{}{}
\makeatother

\makeatletter
\newcommand\blankfootnote[1]{%
  \begingroup
    \renewcommand\@makefnmark{}%
    \renewcommand\@makefntext[1]{\noindent##1}%
    \footnotetext{#1}%
  \endgroup
}
\makeatother

\title[DAG misspecification]{Your Assumed DAG is Wrong\\ and Here's How To Deal With It}
\usepackage{times}
\clearauthor{%
 \Name{Kirtan Padh}\footnote{\label{author} Equal contribution} \Email{kirtan.padh@tum.de}\\
 \Name{Zhufeng Li}\footref{author} \Email{zhufeng.li@helmholtz-munich.de}\\
 \Name{Cecilia Casolo} \Email{cecilia.casolo@helmholtz-munich.de}\\
 \Name{Niki Kilbertus} \Email{niki.kilbertus@tum.de}\\
 \addr Technical University of Munich, Helmholtz Munich, Munich Center for Machine Learning (MCML)
}

\begin{document}

\maketitle

\begin{abstract}%
Assuming a directed acyclic graph (DAG) that represents prior knowledge of causal relationships between variables is a common starting point for cause-effect estimation. Existing literature typically invokes hypothetical domain expert knowledge or causal discovery algorithms to justify this assumption. In practice, neither may propose a single DAG with high confidence. Domain experts are hesitant to rule out dependencies with certainty or have ongoing disputes about relationships; causal discovery often relies on untestable assumptions itself or only provides an equivalence class of DAGs and is commonly sensitive to hyperparameter and threshold choices. We propose an efficient, gradient-based optimization method that provides bounds for causal queries over a collection of causal graphs---compatible with imperfect prior knowledge---that may still be too large for exhaustive enumeration. Our bounds achieve good coverage and sharpness for causal queries such as average treatment effects in linear and non-linear synthetic settings as well as on real-world data. Our approach aims at providing an easy-to-use and widely applicable rebuttal to the valid critique of `What if your assumed DAG is wrong?'.
\end{abstract}


\section{Introduction}
\label{sec:intro}
\blankfootnote{{Code}: \url{https://github.com/zhufengli/your_dag_is_wrong}}
Estimating the strength of causal effects is crucial across numerous domains, as it enables informed decision-making and policy interventions. Causal inference techniques have been applied in a wide variety of fields including but not limited to economics \citep{imbens2009recent,athey2017state}, education \citep{lalonde1986evaluating,sewell1968social,dehejia1998causal}, healthcare \citep{badri2009healthcare,sanchez2022causal}, biology \citep{lecca2021machine}, medicine \citep{castro2020causality,liu2022medical,feuerriegel2024causal}, and software engineering \citep{siebert2023applications}.

The gold standard for quantifying causal relationships is via active experimentation and intervention, such as in (controlled) clinical trials. However, in many settings, such experiments are unethical or impossible. For instance, randomly assigning smoking habits to assess the effect on lung cancer risk would be highly unethical. Similarly, questions like whether and how strongly the health of institutions affects economic growth cannot feasibly be tackled by active experimentation \citep{acemoglu2001colonial}. These challenges, combined with the rapid development of computational data analysis methods in the last few decades, have led to significant advancements in techniques to estimate causal effects from observational data \citep{guo2020survey,malinsky2018causal,peters2017elements,glymour2019review}.

Within many causal inference frameworks, estimating the strength of causal effects requires knowledge of the underlying causal graph. Specifically, the Structural Causal Model (SCM) \citep{pearl2009causal,peters2017elements} framework encodes causal assumptions in the form of Directed Acyclic Graphs (DAGs), where edges represent the data-generating mechanisms in terms of which variables influence others \citep{pearl1995causal,lauritzen2001causal}. This information, together with observational and/or experimental data, can then be used for downstream cause-effect estimation tasks.
However, in practice, the causal graph is often not known with certainty. Most literature on cause-effect estimation assumes that the causal graph is given, typically referring to domain expert knowledge or causal discovery algorithms to justify this assumption. Yet, it is unlikely that even domain experts would assign high confidence to a single graph; they may be more inclined to debate the absence or presence of specific edges rather than agree on a complete structure.

Causal discovery algorithms aim to learn the causal structure from data \citep{spirtes2000causation,spirtes1995causal,zheng2018dags,chickering2002optimal} (see \citet{vowels2022d} for a more comprehensive review). Despite tremendous efforts, causal discovery has not advanced to a point where it is commonly relied upon in practice. So-called `constraint-based' algorithms rely on conditional independence tests (with high-dimensional conditional sets) \citep{spirtes1995causal,spirtes2000causation}. The difficulty of conditional independence testing \citep{Shah_2020,lundborg2022conditional} and adaptively compounding errors often render these methods unreliable in practice. Even with a conditional independence oracle, they can only identify causal graphs up to a `Markov equivalence', i.e., their output is typically a collection of DAGs instead of a single one. Other methods for causal discovery critically rely on other untestable assumptions, such as specific function classes or distributional assumptions \citep{peters2014causal}. So-called `score-based methods', the third major class of causal discovery algorithms typically require solving non-convex constrained optimization problems, and the proposed DAG critically depends on heuristic hyperparameter choices \citep{zheng2018dags,wei2020dags}.
Finally, more recent methods acknowledge that committing to a single DAG is overly restrictive and learn entire distributions over plausible graphs (see \citealp{mamaghan2024challenges} for an overview).
Ultimately, all these methods are practically limited in that there remains room for uncertainties in the final output \citep{reisach2021beware,ijcai2024p374,sondhi2019reduced}.

As a result, the assumed causal graph is an elusive object, and there is often ambiguity about various edges. Estimating causal effects based on a single assumed graph is, therefore, likely equally faulty. In this work, we address the uncertainty inherent in the causal graph structure, particularly concerning the existence and direction of certain edges. This uncertainty could arise from under-performing structure learning algorithms or limited domain knowledge. For instance, in large graphs, uncertainty may manifest in the form of cluster DAGs \citep{anand2023causal}, where there is knowledge about clusters of nodes and inter-cluster relationships, but no information on intra-cluster causal connections. Alternatively, domain experts might be confident about the existence or non-existence of only certain edges in the graph.
We propose an efficient, gradient-based optimization method that computes bounds on cause-effect estimates over all plausible DAGs compatible with the available prior knowledge. Our method is applicable even when the collection of compatible graphs is too large for exhaustive enumeration. Through this approach, we aim to provide a robust framework for causal inference that acknowledges the uncertainty in the assumed causal structure, addressing the critical concern of ``What if your assumed DAG is wrong?''.
Our contributions are as follows:
\begin{itemize}[leftmargin=*,itemsep=-2mm,topsep=0pt]
    \item We develop a method to efficiently bound causal effects over a (potentially large) flexibly pre-defined set of plausible causal graphs.
    \item Our method is applicable in non-linear and continuous settings and for large graphs.
    \item Beyond empirical evaluation on synthetic data, we demonstrate our method on a real-world dataset, where we attempt to find plausible graphs via constraint-based causal discovery.
\end{itemize}

\section{Problem setting and assumptions}
\label{sec:problem-setting}

\subsection{Background and setup}

We operate within the fully observed structural causal model (SCM) framework \citep{pearl2009causality,peters2017elements}, where an SCM is a tuple $(\bV, \cF, N, P_{\N})$.
Here, $\bV = \{X_1, \ldots, X_d\}$ are the endogenous variables, $\N = (N_1, \ldots, N_d)$ are exogenous (or noise) variables, $P_{\N}$ is a distribution over $\N$ where the $\N$ are jointly independent, and $\cF = \{f_i\}_{i=1}^d$ is a collection of functions (the structural equations) such that $X_i := f_i(\paG(X_i), N_i)$ for all $i \in \{1, \ldots, d\}$.
The set $\paG(X_i)\subset \bV$, called `the parents of $X_i$,' is such that the graph $\cG(\bV, \bE)$ over $\bV$ is acyclic.
Here, $\bE \subset \{(X_i, X_j) \in \bV \times \bV \given i \ne j\}$ is the set of edges.
We often denote the DAG induced by the SCM simply by $\cG$ when not explicitly referring to the vertices/edges, write $\pa(X_i)$ when the graph is clear from the context, and use $X_i \to X_j$ for $(X_i, X_j) \in \bE$.
Using Pearl's `do notation' \citep{pearl2009causal}, $do(X_i = x^*)$ denotes a (hard) intervention on $X_i \in \bV$ fixing its value to $x^*$.
Technically, this means replacing $f_i \in \cF$ with the function $X_i := x^*$.
Any fully observed SCM as defined above induces a unique joint distribution $P_{\bV}$ over $\bV$, called `observational distribution,' which is Markov with respect to the graph $\cG$, i.e., each $X_i \in \bV$ is conditionally independent of its non-descendants given its parents \citep{peters2017elements}.\footnote{We assume throughout that densities exist and all random variables have finite variance. Under these mild assumptions, the stated \emph{local} Markov property is equivalent to the observational distribution factorizing according to $\cG$ as well as to the global Markov property, namely that d-separations in $\cG$ imply conditional independencies in the observational distribution \citep{peters2017elements}.}
We will also represent a graph $\cG(\bV, \bE)$ via its adjacency matrix $A_{\cG}\in \{0,1\}^{d \times d}$, where $A_{ij} = 1$ if and only if $(i, j) \in \bE$.
Finally, let $\cD = \{x^i_1, x^i_2, \ldots, x^i_d \}_{i=1}^N$ be a dataset of $N$ i.i.d.\ samples from the observational distribution.
In short, we assume data to be generated by a fully observed SCM, but a priori make no assumptions on the distribution $P_{\N}$ or the functional relationships $\cF$.
In particular, we will consider linear and non-linear $\cF$ separately later.

Let $\cE_{\bV}$ be the set of all possible edge sets that form a valid DAG with vertices $\bV$. The set of allowed DAGs can be noted as $\cG(\bV, \cE)$. 
The setting we are interested in is when we have partial information about causal relations between some of the variables, i.e., sure edges $\bE_{s}$ and forbidden edges $\bE_{f}$. $\bE_{s}$ are edges whose presence we are certain about, and $\bE_{f}$ are edges that we are certain are not present. Let $\cE_{s} \subseteq \cE_{\bV}$ denote the set of edge sets that contain edges in $\bE_{s}$. Similarly, let $\cE_{\neg f} \subseteq \cE_{\bV}$ denote the set of edge sets that do not contain $\bE_{f}$.
With this information given, the set of possible DAGs is reduced to $\cG(\bV, \cE_{s} \cap \cE_{\neg f})$. 

\xhdr{Goal:} Our goal is to estimate bounds on a causal query $\cQ_{\cG}$ based on observational data $\cD$, over all possible graphs $\cG(\bV, \cE_{s} \cap \cE_{\neg f})$. For instance, common natural queries include interventional expectations of the form $\cQ_{\cG} = \E[Y \given do(X = x^*)]$ for $X, Y \in \bV$. A common way to identify interventional probabilities is via the adjustment formula, where given a valid adjustment set $\Z \subset \bV$, we have (assuming the existence of densities throughout)
\begin{equation}
\label{eq:adjustment}
   p\bigl(y \given do(X = x^*)\bigr) = \int p(y \given x^*, \z) p(\z)\, d\z\:.
\end{equation}
This adjustment formula expresses$\cQ_G$ only in terms of the observational distribution, i.e., only relying on observed marginal and conditional densities~\citep{pearl2009causality} as described in \cref{subsec:query}. In the fully observed setting, every interventional distribution is identified from the observational distribution when the ground truth graph $\cG(\bV, \bE)$ is known exactly. However, when the graph is not known with certainty we typically cannot estimate $\cQ_{\cG}$ consistently. Yet, we can obtain lower and upper bounds for the quantity over the set of allowed DAGs. There is a trivial brute-force approach, which is to estimate the target query for every possible graph and take the minimum (maximum) over all these values. Since the space of DAGs grows quickly as the number of nodes increases, this quickly becomes computationally infeasible. Instead, we aim to use continuous optimization in a targeted search for the bounds, i.e., the smallest (largest) possible values of the causal query of interest over the space of DAGs.
We note that while the underlying quantity is a bound set rather than an interval, we only report the minimum and maximum values from this set. 

\subsection{Related Work}
\label{sec:rel-work}

Point identification of causal effects generally requires strong assumptions on the causal structure or the functional relationships between the variables. When these assumptions are not met, effects can often still be \emph{partially identifiable}, meaning that we can confine the effect to lie within some non-empty, but constrained \emph{set} of effects that are compatible with the structural assumptions and the observed data \citep{manski1990nonparametric}. Such effect sets are often characterized by intervals, defined by a lower and upper bound on the target quantity. Initial work on this was pioneered by \citet{Balke1997,chickering96}, and the topic has attracted much attention lately. 

Most of this work has focused on partial identification or sensitivity analysis in the presence of confounding. Different streams of work have focused on bounding causal effects under a variety of assumptions, including assuming discrete domains~\citep{zhang2021bounding, duarte2023automated,raichev2024estimating}, assuming the presence of an instrument~\citep{gunsilius2019path, kilbertus2020class}, using neural networks~\citep{hu2021generative, padh2023stochastic}, or performing causal sensitivity analysis~\citep{frauen2024sharp, melnychuk2024partial, jesson2021quantifying, marmarelis2023partial}.
Other sources of uncertainty in causal inference include ones arising from interaction patterns in non-IID data~\citep{zhang2022causalinferenceuncertainty,bhattacharya2020causal}, shifts in covariates~\citep{Jesson2020identifying}, and unobserved confounders~\citep{Tchetgen2013control}.

Despite these efforts, the uncertainty related to the causal graph itself, particularly in the final goal of treatment effect estimation, has received less attention.
Some works focus on improving the inference under the uncertainty introduced by previous causal discoveries on the same data \citep{chang2024post, gradu2022valid, malinsky2024cautious}.
For constraint-based causal discovery, these are often characterized by Markov equivalence class (MEC) over which one can perform partial identification \citep{maathuis2009estimating, bellot2024towards}.
However, the prevailing approach---focusing on uncertain orientations of unoriented edges in the completed partially directed acyclic graph (CPDAG) corresponding to the MEC---is limited for two primary reasons.
First, adopting a more flexible approach to managing uncertainty in causal graphs could prove beneficial, especially when leveraging domain knowledge to infer edge information.
Second, several studies have highlighted the potentially unreliable performance of (constraint-based) causal discovery algorithms in real-world settings \citep{reisach2021beware}. Hence, uncertainties beyond the ones arising purely due to causal discovery should be considered.
In orthogonal work, \citet{henckel2024adjustment} highlight that the ``utility'' of a specific assumed graph typically depends on the downstream task the graph is used for. Starting from a downstream task, i.e., cause-effect estimation, they measure how `close any given DAG is to the ground truth DAG' by comparing how many correct cause-effect estimands the proposed DAG recovers. This idea leads to novel metrics to measure the quality of causal discovery \emph{with a downstream task in mind}.

More recent work by \citet{strieder2023confidence,strieder2024dual} has developed confidence regions for total causal effects that account for uncertainties in both the causal structure and numerical size of nonzero effects, but their applicability is restricted to linear models.
Additionally, efforts to mitigate uncertainties from imperfect expert input have utilized Large Language Models \citep{long2023causal} and pre-processing techniques \citep{oates2017repair}.

Our work is complementary to these efforts and aims to further fill the gap in inference under the uncertainty coming from the specification of the causal graph. Unlike previous efforts, our method allows a pre-defined set of plausible causal graphs that are more general than the MEC, does not require the linearity assumption, and uses continuous optimization to allow the method to extend more easily to larger graphs. We describe our method in the next section.

\section{General optimization formulation}
\label{sec:non-param}

This section provides a high-level overview of our approach, followed by a more detailed description of the implementation.
With the notation introduced in \cref{sec:problem-setting}, we can define the unknown edge set as $\bE_u$ and assume its cardinality to be $K \in \bN$. We denote by $A_{\alpha}$ a $d\times d$ matrix where the entries corresponding to the uncertain edges in the adjacency matrix $A$ are replaced by $\{\alpha_1, \alpha_2, \ldots \alpha_K\}$.

\begin{figure}
    \centering
    \subfigure[]{
        \includegraphics[width=0.15\textwidth]{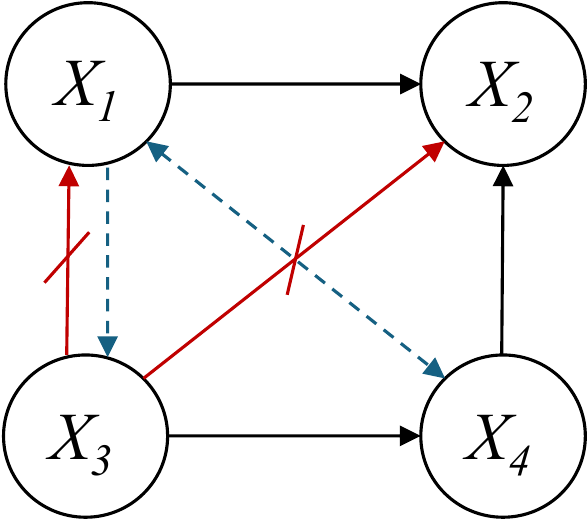}
        \label{fig:example}
    }
    \subfigure[]{
        \includegraphics[width=0.14\textwidth]{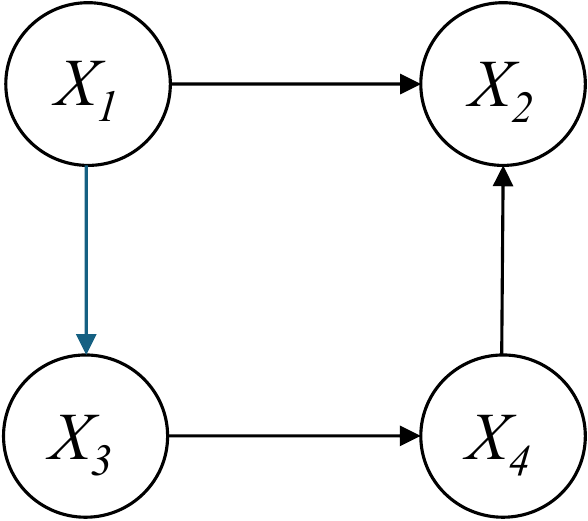}
        \label{fig:plausible1}
    }
    \subfigure[]{
        \includegraphics[width=0.14\textwidth]{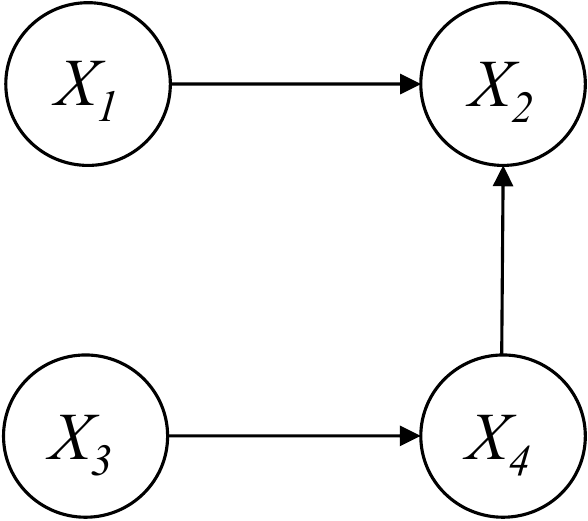}
        \label{fig:plausible2}
    }
    \subfigure[]{
        \includegraphics[width=0.14\textwidth]{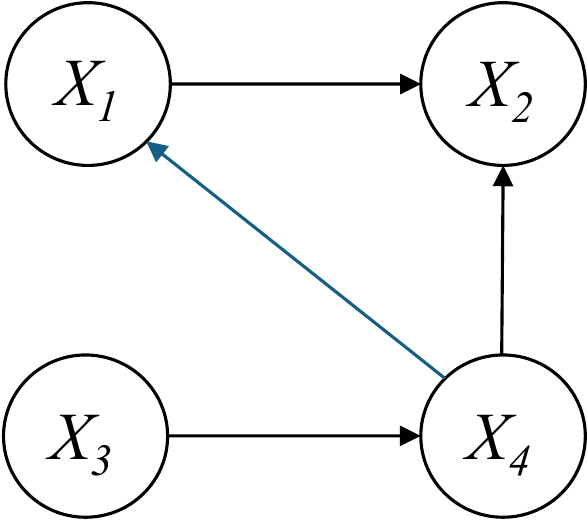}
        \label{fig:plausible3}
    }
    \subfigure[]{
        \includegraphics[width=0.14\textwidth]{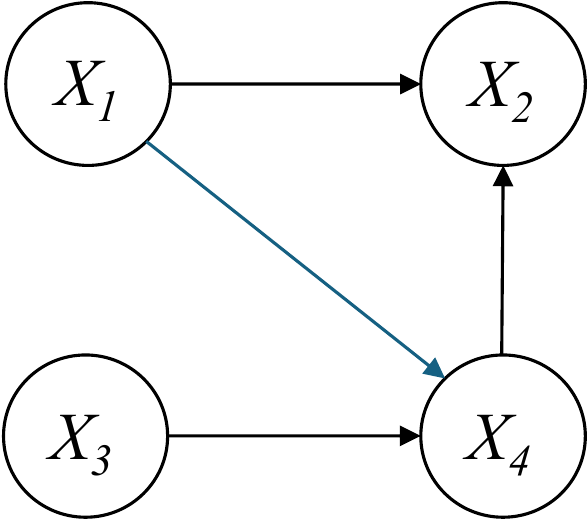}
        \label{fig:plausible4}
    }
    \subfigure[]{
        \includegraphics[width=0.14\textwidth]{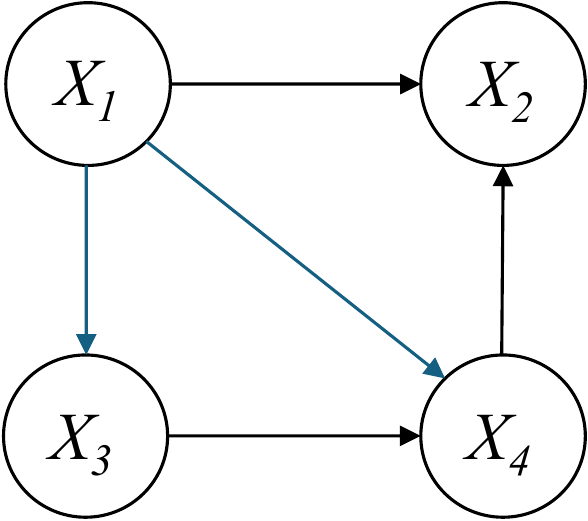}
        \label{fig:plausible5}
    }
    \caption{(a) An example illustrating a graph with partial information. Red edges represents forbidden edges and blue dotted edges represent unknown edges. (b)–(f) All plausible DAGs compatible with the information provided in (a).}
    \label{fig:example-graphs}
\end{figure}

\subsection{Illustrative example}

\Cref{fig:example} shows an example, where the sure edge set is $\bE_s = \{(X_1,X_2), (X_3,X_4), (X_4, X_2)\}$ and the forbidden edge set is $\bE_f = \{(X_3, X_1), (X_3, X_2)\}$. With the above knowledge, we are still unsure about the existence of the edge between $X_1$ and $X_3$, as well as between $X_1$ and $X_4$, even in terms of their direction.
Therefore, we have $\bE_u = \{(X_1, X_3), (X_1, X_4), (X_4, X_1), \emptyset\}$ and thus $K = 3$.
Assuming the quantity $\cQ$ we are interested in is $\E[X_2 \given do(x^*_3)]$, computing $\cQ$ depends on the (non)existence of these unknown edges.
For the same observed data, we, therefore, have more than one possible value for $\cQ$ given the different possible graph structures shown in \cref{fig:plausible1,fig:plausible2,fig:plausible3,fig:plausible4,fig:plausible5}.
The minimum and maximum of these values form our bounds.
Estimating $\cQ$ could involve estimating and combining conditional and marginal densities from observed data in the general case.
However, cause effect estimation can often be phrased as a combination rather simple `function fitting' steps in specific cases, as we will describe in \cref{subsec:estimating-q}. More details on further possible causal queries can be found in \cref{app:queries}.
The following matrix represents the adjacency matrix for the graph shown in \cref{fig:example} including its uncertain entries (we use row/column indices for the $\alpha$s instead of a single running index for readability) 
\begin{equation}
\label{eq:mat-ex}
  \text{$A_\alpha$} = \kbordermatrix{
    & X_1 & X_2 & X_3  & X_4\\
    X_1 & 0 & 1 & \alpha_{13} & \alpha_{14}  \\
    X_2 & 0 & 0 & 0 & 0  \\
    X_3 & 0 & 0 & 0 & 1  \\
    X_4 & \alpha_{41} & 1 & 0 & 0 
  }\:.
\end{equation}
Not all combinations of edges in $\bE_u$ form a plausible graph because of the acyclicity constraint.
In this example, our search space $\cE_s \cap \cE_{\neg f} = \bE_s \cup \{\{(X_1,X_3),(X_1,X_4)\}, \{(X_1,X_3)\}, \{(X_1, X_4)\}, \allowbreak \{(X_4,X_1)\}, \{\emptyset\} \}$ is small enough for the brute force approach, but we will show how to get these bounds using continuous optimization.
Let $\hat{\cQ}$ be the estimated $\cQ$ from observed data.
We are then looking at the following optimization problem to obtain the lower / upper bounds (for $K=3$)
\begin{equation}
\label{eq:gen}
\min\limits_{\alpha \in {\{0, 1\}}^{K}} / \max\limits_{\alpha \in {\{0, 1\}}^{K}} \hat{\cQ}(A_{\alpha}, \cD)\:, \qquad
\text { subject to } \cG(A_{\alpha}) \in \mathrm{DAGs}\:,
\end{equation}
where we `flatten' the variable edges of $A_{\alpha}$ into a vector $\alpha = (\alpha_{13}, \alpha_{14}, \alpha_{41}) \in \{0,1\}^K$ in arbitrary order.
In words, find the minimum and maximum estimates of $\cQ$ over all values of $\alpha$ for which $A_{\alpha}$ represents a DAG.
We note that the search space grows super-exponentially in $K$, the number of uncertain edges.
Due to the combinatorial nature of the problem, global optimization that is more efficient than a brute force search is not straight forward.
We tackle this challenge by leveraging similarities to the optimization formulation underlying score-based continuous causal discovery methods.
In particular, we show how to build on concepts developed by \citet{zheng2018dags} to convert the constrained discrete optimization into an unconstrained continuous optimization problem for which we can use augmented Lagrangian techniques combined with differentiable DAG sampling approaches proposed by, e.g., \citet{charpentier2021differentiable}, which have also been updated to incorporate prior knowledge \citep{rittel2023specifying}.

\begin{figure}
    \centering
    \includegraphics[width=0.9\textwidth]{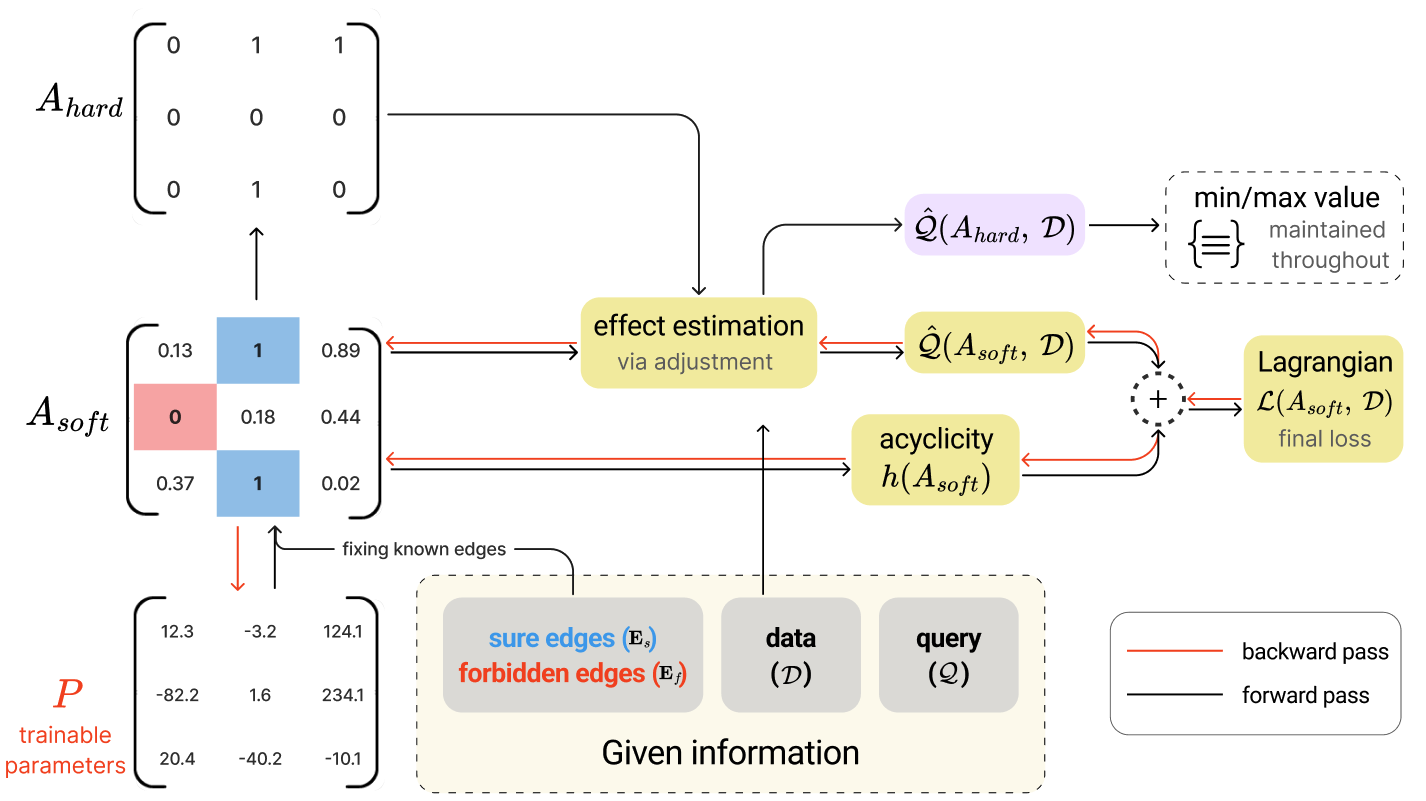}
    \caption{An illustrative diagram of the Lagrangian method.} 
    \label{fig:method-figure}
\end{figure}

\subsection{Solving the general optimization problem}
\label{subsec:solve-gen}
To operationalize this optimization, we first re-formulate \cref{eq:gen} as a continuous unconstrained optimization problem without introducing any parametric assumptions. 
\Cref{fig:method-figure} illustrates our overall framework. In this subsection and \cref{sec:solve-opt}, we motivate and describe all aspects of this diagram one by one.
The first step is to provide an estimator of the causal query $\cQ$.

\subsubsection{Estimating \texorpdfstring{$\cQ$}{Q}}
\label{subsec:estimating-q}

We resort to cause-effect estimation via valid adjustment sets \citep{perkovic2015complete, perkovic2018complete, shpitser2010adjustment} to control for confounding variables.
Concretely, let $X, Y \in \bV$, where $X$ is the `treatment' and $Y$ is the `outcome'.
One of the primary challenges in cause-effect estimation comes from confounding variables that influence both the treatment and the outcome. By `controlling' or `adjusting' for these variables, the adjustment set isolates the direct causal effect of the treatment on the outcome from the overall association, which may be due to direct, but also confounding effects.
A set $\Z \subset \bV$ is called valid adjustment set for $X,Y$ in the DAG $\cG(\bV, \bE)$ if (i) $\Z$ blocks all undirected paths in $\cG$ between $X$ and $Y$, and (ii) $\Z$ does not contain any node on a directed path from $X$ to $Y$ in $\cG$ or any descendant of such a node \citep{shpitser2010adjustment}.
There may be multiple valid adjustment sets for a tuple $X,Y$ and we implement our method using both parent and optimal adjustment sets to estimate the query \citep{Henckel_2022}.
Selecting these adjustment sets without interfering with the gradient-based continuous optimization involves some non-trivial technicalities.
We provide further details on adjustment sets and pseudocode on how we adapt them to fit into our framework in \cref{app:adjustment-sets}.
Given a valid adjustment set for $X,Y$ in $\cG$ allows us to estimate the causal effect of $X$ on $Y$, i.e., the interventional distribution, via \cref{eq:adjustment}.
Naive estimation of this estimand would require conditional density estimation of $p(y \given x, \z)$ and an empirical (e.g., sampling-based) integration over the marginal $p(\z)$.
In general, the marginal density could be estimated via a broad range of methods from Gaussian processes, to mixture density network \citep{bishop1994mixture} or conditional normalizing flows/diffusion/flow matching \citep{papamakarios2021,ho2020denoising,lipman2022flow}.
Common targeted causal queries such as the Average Treatment Effect (ATE) and the Conditional Average Treatment Effect (CATE) can be estimated directly from estimates of the general interventional distribution in \cref{eq:adjustment}.
Depending on further parametric assumptions, conditional density estimation can be circumvented and expected causal effect estimation can be reduced more effectively to simple function fitting routines, see \cref{subsec:query}.

\subsubsection{Enforcing acyclicity}

\citet{zheng2018dags} are widely credited for the introduction of a smooth function of weighted (continuous) adjacency matrices, $h: \bR^{d\times d} \to \bR_{\ge 0}, \: A\mapsto \tr(\exp(A \odot A)) - d$ (where $\odot$ denotes the Hadamard product) that captures the `degree of acyclicity' of the corresponding graph.
This function has several desirable properties:
(i) It quantifies the degree to which a graph is acyclic referred to as `DAG-ness' (if $h(A) = 0$, $A$ represents an acyclic graph; the larger $h(A)$ the more `cyclic' the graph). (ii) It is smooth. (iii) Both the function and its derivative are computationally tractable.
By replacing the combinatorial acyclicity constraint ($\cG(A_{\alpha} \in \mathrm{DAGs}$)  with the equality constraint $h(A_{\alpha}) = 0$, the DAG learning problem becomes amenable to conventional continuous constrained optimization techniques based on gradient descent.
This enables us to leverage the robust and efficient optimization tools provided by modern machine learning frameworks.
Since the original proposal of $h$ by \citet{zheng2018dags}, various other constraints have been proposed that offer computational or numerical benefits. \citet[Sec.~5]{vowels2022d} provide an incomplete list.
In this work we use the constraint defined by \cite{bello2022dagma} which performs well empirically.

\subsubsection{Putting it all together}

With the adjustment based estimand and a continuous equality constraint for acyclicity, what remains is to convert the discrete search space into a continuous one, in which the the query is continuous (and differentiable).
Specifically, we would like to replace the search space $\{0, 1\}^K$ in \cref{eq:gen} with $\bR^{K}$.
A direct relaxation does not work, as we require strict zeros to indicate the missingness of edges, which in turn affects which $\Z$ form valid adjustment sets for a given tuple $X,Y$.
This challenge can be addressed via the Gumble-Softmax straight-through estimator \citep{jang2016categorical}, which allows us to include discrete distributions as part of an overall differentiable optimization procedure.
We describe the details in \cref{subsec:gumble}.
Finally, putting all these building blocks together allows for a re-formulation of the resulting constrained continuous optimization problem into an unconstrained one using the augmented Lagrangian formulation \citep{birgin2014practical} as detailed in \cref{subsec:lagrangian}.

We should point out that this is far from the only way to tackle this optimization. Any of the array of score-based continuous causal discovery methods~\citep{wei2020dags, montagna2023scalable, bello2022dagma, charpentier2021differentiable, zheng2018dags} could be used as well if adapted correctly, and we hope that this work serves as a starting point to do so, as well as provide motivation to take into account the final quantity of interest already at the time of constructing the graph. In particular, in the next sections, we focus on \citet{charpentier2021differentiable} and \citet{zheng2018dags}.

\section{Operationalizing the Optimization}
\label{sec:solve-opt}

\subsection{Discrete distributions: Straight-Through Gumbel-Softmax (ST-GS) estimator}
\label{subsec:gumble}

Representing a directed graph numerically via a binary adjacency matrix is a natural choice.
However, any function of this binary matrix will be non-differentiable and thus forestall the use of backpropagation.
To allow gradients to flow through the binary adjacency matrix, required to optimize over graphs, we use the Gumbel-Softmax distribution relaxation \citep{jang2016categorical}, which approximates categorical samples from continuous variables via a simple Softmax calculation.
At a high level, the idea is to bridge the gap between a continuous parameter set and the discrete search space through the use of the Gumble-Softmax distribution. It is called a Gumbel-Softmax straight-through estimator because the forward pass uses a discrete, binary sampled matrix, while the backward pass employs the softmax probabilities derived from the same logits, ensuring differentiability despite the discretization in the forward pass.
Following \citet{rittel2023specifying}, in practice, we optimize over a matrix $P\in \bR^{d\times d}$ (see \cref{fig:method-figure}), in which some entries get frozen (sure/forbidden edges) to obtain a `soft` adjacency matrix $A_{soft}$. The soft adjacency matrix parameterizes a Gumble-Softmax distribution from which a `hard` binary adjacency matrix $A_{hard}$ is sampled to represent the actual causal graph $\cG$ used in the cause-effect estimation of the forward pass, e.g., to find valid adjustment sets.
In the backward pass, the gradients flow through the `soft` adjacency matrix to the continuous parameters actually being optimized.
Details are described in \cref{app:gumble}.
For the actual optimization over DAGs we focus on the (augmented) Lagrangian \citep{zheng2018dags} and the DP-DAG \citep{charpentier2021differentiable} methods.

\subsection{Optimizing over the space of DAGs}
\label{subsec:lagrangian}

\xhdr{Lagrangian optimization.} Denoting our estimator of the causal query $\cQ$ by $\hat{\cQ}$ and employing the acyclicity constraint as described in \cref{subsec:solve-gen}, the optimization problem in \cref{eq:gen} becomes
\begin{equation}
\label{eq:opt-gen}
\min\limits_{\alpha \in \bR^{K}} / \max\limits_{\alpha \in \bR^{K}}\: \hat{\cQ}(A_{\alpha}, \cD)\:, \qquad
\text { subject to } h(A_{\alpha}) = 0\:.
\end{equation}
We use the augmented Lagrangian method for (in)equality constraints to solve the constrained optimization problem in \cref{eq:opt-gen} by converting it into an unconstrained problem. The formulation is taken from \citet[Sec.~17.3]{nocedal2006numerical}.
Further details are provided in \cref{app:lagrangian}, which includes the pseudocode for implementing the Lagrangian method in both linear and non-linear settings, presented respectively in \cref{alg:linear} and \cref{alg:nonlinear}.

\begin{figure}
    \centering
    \includegraphics[width=0.9\textwidth]{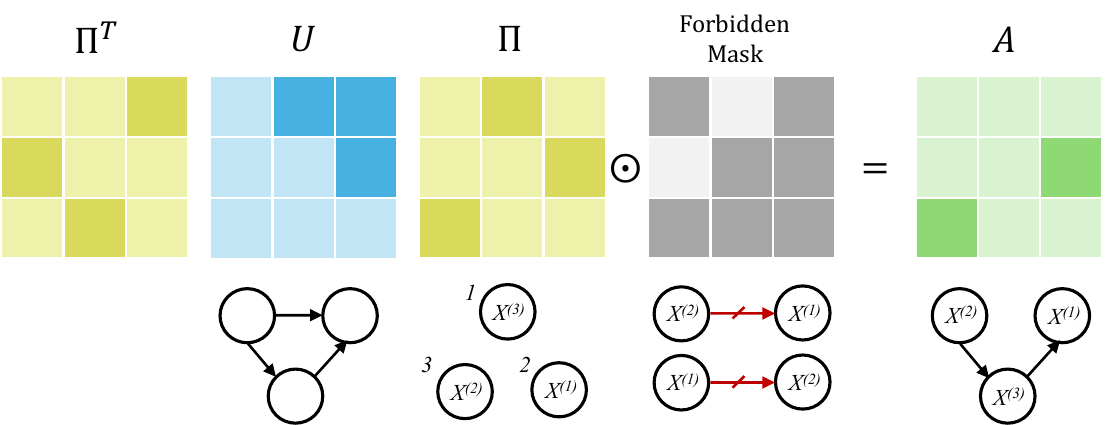}
    \caption{An illustration of the DP-DAG sampling procedure with forbidden edge masking, where dark and light colors represent 1s and 0s, respectively.
    The graphical interpretation of each matrix are shown below.} 
    \label{fig:dpdag-figure} %
\end{figure}

\xhdr{Differentiable DAG sampling.}
In the augmented Lagrangian method, the constraints are enforced essentially via penalty terms that can rarely be ensured to be exactly zero.
Hence, they still sometimes yield cyclic graphs after convergence.
Therefore, we additionally apply the DP-DAG method~\citep{charpentier2021differentiable}, which provides another solution for sampling DAGs from a trainable distribution.
Permitting node indices permutations, any full DAG can be represented as an upper triangular adjacency matrix with zero diagonal and all non-zero entries being $1$. 

DP-DAG samples a DAG by generating an upper triangular matrix $U$, which serves as the graph topology, i.e., a plain graph structure without attribution of node indices.
In order to attribute indices to the graph, it generates a permutation matrix $\Pi$.
The permutation is applied on the row and column indices to preserve the graph skeleton, resulting in $A=\Pi^T U \Pi$.
We illustrate this process in \cref{fig:dpdag-figure}.
Removing edges in the obtained graph simply amounts to replacing 1 entries with 0s via masking. Removing edges can not introduce directed cycles.
However, enforcing sure edges by naively replacing 0 entries with 1s may introduce directed cycles, which puts DP-DAG at a certain disadvantage compared to the augmented Lagrangian method.
Enforcing sure edges shrinks the search space and can thus allow for narrower bounds. However, the bounds obtained from ignoring sure edges and treating them as uncertain instead, will still be valid---albeit potentially looser.
In our experiments, we only enforce forbidden edges when using the DP-DAG approach.

\subsection{Query: Average effects}
\label{subsec:query}

While any query $\cQ$, i.e., any functional of interventional distributions, can generally be estimated from observational data as described in \cref{subsec:estimating-q}, we can significantly simplify the formulation when restricting our attention to the expected value $\cQ = \E[Y \given do(X = x^*)]$ as our query of interest, from which we can directly compute, for example, average treatment effects.
Let $\Z$ be a valid adjustment set for $X, Y$ in a given DAG $\cG$.
Then we have 
\begin{equation}
\label{eq:exp-q}
        \cQ = \int y\, p(y \given do(X = x^*))\, dy
        = \int y \int p(y \given x^*, \z)p(\z)\, d\z\, dy
        = \int p(\z) \E[Y \given x^*, \z]\, d\z\:, %
\end{equation}
which we can estimate directly form observational data via $\hat{\cQ} = \tfrac{1}{N} \sum_{i=1}^N \hat{\E}[Y \given x^*, \z_i]$ for any estimator of the conditional expectation, i.e., any machine learning regressor predicting $Y$ from $x$ and $\z$.
Depending on the dimensionality and modality of $\Z$, we may choose to estimate the density of $\Z$, in which case a Monte Carlo approximation of the integral over $\Z$ can be computed with sampling access to $\Z$. 
We highlight that the nodes in $\Z$, the valid adjustment set, depends on the optimization parameters $\alpha$. Hence, the dependence of the objective $\cQ$ on $\alpha$, representing the graph, comes solely from what constitutes a valid adjustment set for $X, Y$. 
In practice, to evaluate $\cQ(A_{\alpha}, \cD)$, the query for the current value of $\alpha$ and given data $\cD$, we thus only need to fit a single regression function.
Hence, our approach is directly applicable to nonlinear contexts, where we use small Multilayer Perceptron (MLP) to estimate $\E[Y \given x, \z]$.
In linear settings, we can even sidestep the empirical mean over $\z$, by directly reading off the causal effect from the coefficient of $x$ in a single ordinary least squares linear regression from $X, \Z$ to $Y$ \citep{Henckel_2022}.

\section{Experimental results}

In this section, we evaluate our algorithm across different synthetic data scenarios, which differ in their data-generating processes and levels of uncertainty. Additionally, we apply our algorithm to the real-world Infant Health and Development Program (IHDP) dataset \citep{hill2011bayesian}.
Hyperparameter choices and specifics of the data-generating processes are in \cref{app:experiment-setup}. Additional implementation details and information on the packages and data resources used in the implementation can be found in \cref{app:imp-details} and \cref{app:code} respectively.

\begin{figure}[t]
    \centering
    \subfigure[Linear mechanism (1890 simulations)]{ %
        \centering
        \includegraphics[width=0.45\linewidth]{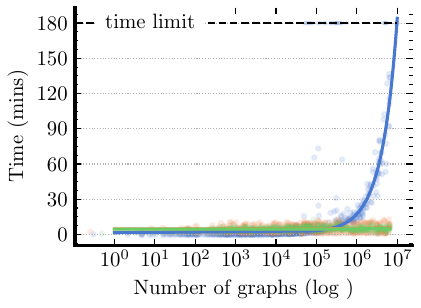}
        \label{fig:runtime_linear}
    }
    \subfigure[Non-linear mechanism (3780 simulations)]{ %
        \centering
        \includegraphics[width=0.45\linewidth]{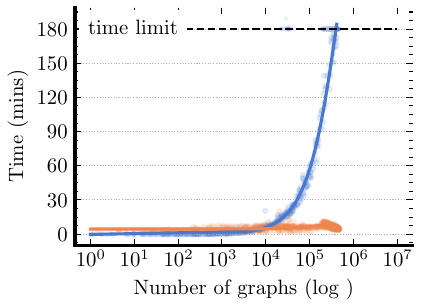}
        \label{fig:runtime_nonlinear}
    }
    \subfigure[Linear mechanism (2322 simulations)]{ %
        \centering
        \includegraphics[width=0.9\linewidth]{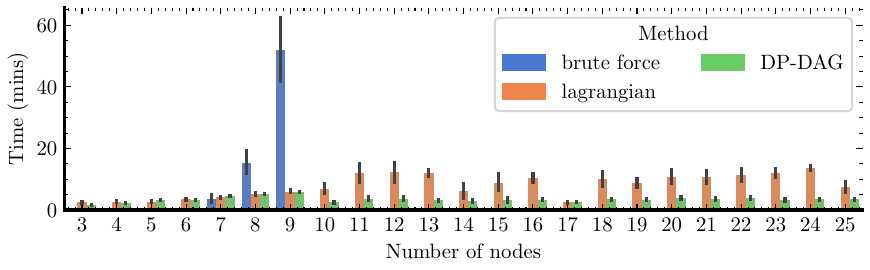}
        \label{fig:runtime_nodes}
    }
    
    \caption{\textbf{Runtime comparison}. Our proposed optimization algorithms scale almost constant over multiple orders of magnitudes in the number of graphs in the search space, compared to a super-exponential growth in runtime for the brute force algorithm. The top row shows the running time as a function of number of possible graphs in the class of misspecification. The bottom row shows the results grouped by the number of nodes. (see \cref{tab:grid-random} for the details of the simulations)}
    \label{fig:runtime}
\end{figure}

\xhdr{Data and uncertainty generation.} We sample random Erdős–Rényi graphs and use one of the three causal mechanisms (linear, sigmoid) described in \cref{tab:mechanism} to generate data based on each graph.
Uncertain edges are either obtained (i) \emph{randomly}, by assigning a pre-set proportion of the true edges and non-edges as known uniformly at random, or (ii) \emph{using the PC-algorithm} for causal discovery \citep{spirtes2000causation}, running it on different permutations of the variables $\bV$ and taking the intersection of directed edges and missing edges across the runs as known and all others as unknown.
The latter option illustrates challenges in the `discovery then estimation' approach to causal inference purely from observational data without any additional assumptions.

\xhdr{Comparison metrics.} Estimated bounds should contain the ground truth effect while being as informative, i.e., narrow, as possible.
Moreover, runtime and scalability is an important aspect in this problem as the reference, brute force search, becomes intractable quickly.
This motivates the following metrics computed individually for each run, denoting true and estimated lower/upper bounds on $\cQ$ by $\lb$/$\ub$ and $\hat{\lb},\hat{\ub}$, respectively (while assuming $\lb\le \ub, \hat{\lb} \le \hat{\ub}$).\footnote{In synthetic settings we can obtain ground truth bounds for small graphs by a brute force search over all DAGs compatible with the prior knowledge of forbidden and sure edges.}
\begin{itemize}[leftmargin=*,itemsep=-2mm,topsep=0pt]
    \item \textbf{Runtime} is the time elapsed for each method on the same hardware.
    \item \textbf{Point coverage} is a binary indicator whether the true causal effect is covered by the estimated bounds, i.e., $1[\hat{\lb} \leq \text{ground truth } \cQ \leq \hat{\ub}]$.
    \item \textbf{Bound coverage} measures the proportion of ground truth bounds that are covered by the estimated bounds and is given by $\frac{|[\lb, \ub] \cap [\hat{\lb}, \hat{\ub}]|}{|[\lb, \ub]|}$, where $|\cdot|$ denotes the length of intervals (Lebesgue measure on $\bR$).
    This metric lies in $[0,1]$ with $0$ meaning no overlap and $1$ meaning that the ground truth bounds are fully contained within the estimated bounds.
    \item \textbf{Bound narrowness} measures how much wider the estimated bounds are compared to the ground truth bounds, which we compute as $\frac{|[\hat{\ub} - \hat{\lb}]|}{|[\lb, \ub] \cap [\hat{\lb}, \hat{\ub}]|}$.
    This metric lies in $[1, \infty)$ where $1$ indicates that the estimated bounds are at least as narrow as the ground truth bounds and larger values mean wider estimated bounds.
    When both bound coverage and narrowness are $1$, we perfectly recover the true bounds.
\end{itemize}

\xhdr{Runtime comparison.} \Cref{fig:runtime} demonstrates that our proposed methods can handle larger, higher-dimensional efficiently with near constant runtime over a wide range of search space sizes. We hypothesize that this near constant scaling comes from the fact that modern gradient-based optimization frameworks (as implemented in PyTorch, jax, etc) are heavily tuned such that an update of $\mathcal{O}(d^2)$ parameters takes roughly equally long for a wide range of $d$.
The brute force approach shows the expected super-exponential increase in computation time with more nodes. Further experiments on runtime comparison are presented in \cref{app:add-experiments}.

\xhdr{Bound coverage/narrowness.} \Cref{tab:metrics-main} shows point coverage and bound coverage, which are both close to $1$ for our approaches. (They are $1$ by definition for the brute force search.)
The bound narrowness reflects the trade off we have to make for the gained efficiency. Since first-order methods (especially over discrete search spaces) may converge to local optima, our bounds are often wider than they need to be---albeit still valid.
Our accounting for finite sample estimation uncertainty in the obtained bounds as discussed in \cref{app:estimation-uncertainty} plays a role in bound narrowness, as we deliberately and conservatively widen the bounds to account for estimation uncertainty.
In cases where the true bounds are narrow, this accounting for finite sample uncertainty can lead to outliers in bound narrowness, substantially influencing the mean values reported here. Bound coverage results for different adjustment set choices and uncertainty sources can be found in \cref{app:add-experiments}.

\begin{table}
\begin{tabular}{lcccccc}
\toprule
        \texttt{Mechanism}   & \multicolumn{2}{c}{\textbf{Point Coverage}} & \multicolumn{2}{c}{\textbf{Bound Coverage}} & \multicolumn{2}{c}{\textbf{Bound Narrowness}} \\ 
           \cmidrule(lr){2-3}\cmidrule(lr){4-5}\cmidrule(lr){6-7}
    \texttt{(n=5670)}       & Lagrangian    & DP-DAG   & Lagrangian   & DP-DAG  & Lagrangian   & DP-DAG       \\ \midrule
\texttt{linear}    & $0.97 \pm 0.0$       & $\mathbf{1 \pm 0.0} $        & $0.93 \pm 0.0$         & $\mathbf{0.99 \pm 0.0}$   & $\mathbf{2.46 \pm 1.35}$   & $2.51 \pm 0.12$          \\
\texttt{non-linear}  & $0.99 \pm 0.0$        & N/A     & $0.93 \pm 0.0$       & N/A       & $2.24 \pm 0.03$         & N/A            \\ \bottomrule
\end{tabular}
\caption{Mean values of our key metrics with standard error across all synthetic data simulations where the uncertain edges are chosen randomly.}
\label{tab:metrics-main}
\end{table}

\subsection{Real-world experiment: IHDP dataset}
\label{subsec:real-world}

In real-world applications, perfect information about the underlying causal graph is rarely available.
In some settings, partial background knowledge about surely existing or definitely non-existing edges may be available, though.
However, in other contexts, domain-specific information may not be entirely relied upon at all.
We now show how one can approach cause-effect estimation in a real-world setting entirely from observational data without relying on any prior information about the causal structure.

A common approach for cause-effect estimation is to run causal discovery as a first stage and use the resulting graph as input for the cause-effect estimation part.
However, causal discovery is an inherently difficult task, known to have multiple limitations \citep{reisach2021beware, ijcai2024p374, sondhi2019reduced}.
When not willing to make additional assumptions about the functional form of the causal relationships, constraint-based causal discovery algorithms such as the PC algorithm \citet{spirtes2000causation} are a common choice. 
However, the PC algorithm can only ever recover an equivalence class, i.e., we are left with a collection of graphs, not a unique one.
Moreover, it presents various limitations: the PC algorithm does not scale well to high dimensions \citet{glymour2019review}, the required conditional independence tests are hard \citep{Shah_2020,lundborg2022conditional}, error propagation is difficult to handle throughout the algorithm \citep{harris2013pc}, and the outcome of the PC algorithm (in its standard form) may even depend on the order in which variables are provided in the input \citep{wienobst2021approach}.
Overall, without strong assumptions, it is challenging to obtain a unique causal graph that can be trusted with high confidence from causal discovery methods.
Instead, we leverage these inherent uncertainties of the PC algorithm output.
We apply the PC algorithm to different permutations of the variables in the dataset, obtaining a Markov equivalence class of graphs each time.
We then interpret edges present in all graphs as sure edges, and the ones always missing as forbidden edges in our framework.

We test this approach on the Infant Health and Development Program (IHDP) dataset \citep{hill2011bayesian}, data from a randomized trial that examined the effects of childcare home visits by specialists on low-birth-weight children's future cognitive test scores.
Selection bias is introduced to the data artificially by removing a subset of the patients, specifically all children with nonwhite mothers, as in \citet{hill2011bayesian, shalit2017estimating}. We follow the data prepossessing of \citet{louizos2017causal}, using the simulated outcomes implemented in the NPCI package \citep{dorie2016npci} and average over 1000 realizations of the outcomes. The analyzed dataset includes 25 covariates about mothers and children among 747 data samples, among which are 139 treated and 608 control. The study collected numerous pre-treatment variables about the children (such as birth information and neonatal health index) and behavior during pregnancy as well as demographic information on the mothers. For further details on the dataset, see \citet{hill2011bayesian}. For causal discovery, we use the PC implementation in \texttt{causallearn} package \citep{zheng2024causal} with the kernel-based conditional independence test from \citet{zheng2011kernel}. We applied the PC algorithm to 10 covariates perturbations.
We apply our nonlinear bounding framework to this dataset and assess the coverage of the ground truth causal effect, as shown in \cref{tab:real-world}.

\begin{table}
\centering
\begin{tabular}{ccc}
\toprule
\textbf{Lower Bound} & \textbf{Upper Bound} & \textbf{Ground Truth Effect} \\ \midrule
$2.4 \pm 2.01$ & $7.2 \pm 2.1$ & $4.7 \pm 0.13$ \\ \bottomrule
\end{tabular}
\caption{The estimated bounds and ground truth effect of the IHDP dataset. The errors are the standard deviations to 10 bootstrap sampling iterations with replacement.}
\label{tab:real-world}
\end{table}

\section{Discussion}

In this paper, we tackled the problem of graph misspecification in causal inference by proposing a method to account for uncertainties in the knowledge of the causal graph. Our approach addresses an essential and common critique of DAG-based causal methods: ``What if the assumed causal graph is incorrect?'' Our method supports more robust conclusions even when the exact causal structure remains uncertain by offering a systematic and computationally efficient way to compute bounds on causal effects across a set of plausible graphs.

Moreover, we provided a practical recipe for integrating our approach with causal discovery algorithms, allowing practitioners to manage uncertainties in a more methodical way rather than relying on a single estimated graph. This approach facilitates capturing the variability in causal estimates that may arise from alternative plausible graph structures, adding a layer of reliability to causal analysis. Our experimental results highlight that the method delivers meaningful bounds on causal effects across various graph configurations, including both linear and non-linear settings, and simulated and real-world cases, in a computationally tractable manner.

\textbf{Limitations:} Our method is limited to the fully observed setting where there is no hidden confounding, limiting its applicability in practical settings where there is virtually always confounding. The method also involves formulating a non-convex relaxation of a discrete search space. Due to the nature of the optimization, we do not have a guarantee that we can get optimal bounds through our methods. In addition, the Lagrangian is not guaranteed to search only over DAGs. While we attempt to overcome this through a filter in the optimization process, it remains a general problem of Lagrangian-based DAG search that it is not guaranteed to return a DAG. The assumption that the relationship between variables forms a DAG is also implicit in this work and has been often challenged in literature.

\textbf{Future directions}: There are several key areas for potential extensions of this work. First, the method could be expanded to address unobserved confounding by incorporating hidden variables within the graph structure. Accounting for graphical structures more general or different than a DAG would also help to improve the applicability of the method. Another possible direction would involve adapting the approach to work with methods beyond the current Lagrangian and DP-DAG techniques; each such adaptation would likely require non-trivial modifications to handle sure and forbidden edges effectively. Additionally, future work could explore defining and optimizing over alternative representations of uncertainty.

In sum, our approach provides a valuable tool for addressing graph uncertainty in causal inference and lays the groundwork for future improvements and applications.

\section*{Acknowledgments}
Cecilia Casolo is supported by the DAAD programme Konrad Zuse Schools of Excellence in Artificial Intelligence, sponsored by the Federal Ministry of Education and Research. 

\bibliography{bib.bib}

\clearpage

\appendix

\section{Possible causal queries}
\label{app:queries}

While we focus on Average Treatment Effect (ATE), different causal queries could be bounded using this infrastructure. As seen before, the do-notation is employed to describe specific interventional scenarios within a causal model. For instance, the query $ p(y \mid do(X = x)) $ represents the distribution of the outcome $ Y $ following an intervention where $ X $ is set to a specific value $ x $, isolating the direct causal effect of $ X $ on $ Y $. To assess the impact of varying levels of intervention, comparative queries such as $ p(y \mid do(X = x_1)) $ versus $ p(y \mid do(X = x_2)) $ are used. Although we focus on single-variable causal effect estimation, a straightforward application of our framework is when the cause and effect are multiple variables, i.e., $\Y = \{Y_1, Y_2, \ldots, Y_{d_Y}\}$ and $\X = \{X_1, X_2, \ldots, X_{d_X}\}$, then the query takes the form of  $ \cQ = p(\y \mid do(\X = \x)) $. Additionally, to allow for heterogeneity, queries like $ p(Y \mid do(X = x), \Z = \z) $ condition on other variables $ \Z $ (while we use the same symbol, this does not refer to a valid adjustment set here) to look at causal effects within different subgroups defined by $\Z = \z$. The calculation of differences in expected outcomes, $ E[Y \mid do(X = x_1)] - E[Y \mid do(X = x_2)] $, quantifies the effect size of an intervention, typically denoted by the Average Treatment Effect (ATE). Lastly, longitudinal studies might use $ p(y \mid do(X_1 = x_1, X_2 = x_2, \ldots, X_k = x_k)) $ to understand the effects of sequential interventions over time. 

\section{Experiment setup}
\label{app:experiment-setup}

\subsection{Data generation}
We follow \citet{Brouillard2020differentiable} in defining the data-generating process in the following steps: 
1. We randomly order the nodes to establish a topological ordering. 2. Edges are chosen according to Erdős–Rényi model while ensuring the causal ordering from step 1. Each potential edge is included in the graph independently with probability $p$, which controls the density of the graph. 3. For each node, the functional form of its causal mechanism is chosen to be among i)linear additive model ii)sigmoidal mixed model, or iii)sigmoidal additive model. The corresponding parameters are then sampled uniformly from preset ranges. 4. Observations are generated in topological order (from root nodes to leaf nodes). Root node values are drawn from a uniform initial distribution, with Gaussian noise added at a coefficient of $0.1$. Each generated observational dataset consists of $5,000$ samples. We summarize the causal mechanisms we are using along with the preset parameter ranges in table \cref{tab:mechanism}.

\subsection{Defining the uncertainty}
\label{app:graph-uncertainty}
Once a ground truth graph and the associated data are generated as described above, we get sure and forbidden edges in two different ways.
\begin{itemize}[leftmargin=*,itemsep=-1.5mm]
    \item \textbf{Random choice:} We set a \texttt{sure edge probability} that represents the proportion of the edges present in the ground truth graph to be included in the 'sure edges' set and a \texttt{forbidden edge probability} that represents the proportion of the edges absent from the ground truth graph to be included in the 'forbidden edges' set.
    \item \textbf{PC algorithm:} We run the PC algorithm with Fisher's test for a number of permutations on the order of the input variables. Each permutation does not usually give the same result as the order in which conditional independence tests are performed in the PC algorithm matters for the final outcome. We take the set of sure edges to be edges that exist in each of the graphs and the set of forbidden edges to be edges that do not exist in any of the graphs. It is crucial to note here that due to the way the sure and forbidden edges are generated in this setting, the set of possible graphs we get from this method might not contain the ground truth graph that generated the data. This is because there might be an edge in the true graph which is always removed by the PC algorithm or vice versa for a non-existing edge
\end{itemize}

\begin{table*}
\centering 
\caption{Data-generating models that we have used in the paper with all parameter ranges}\label{tab:mechanism} 
\begin{tabular*}{\textwidth}{@{\extracolsep{\fill}} l l l}

\toprule

\textbf{Model Name} & \textbf{Formulation} & \textbf{Parameters Range} \\

\midrule
linear additive model    &  $y=\sum_{i\in {\mathrm{pa}}}\beta_{i}x_{i} + \lambda\epsilon$     &  $ \beta \sim \mathrm{Uniform([-1, -0.25] \cup [0.25, 1])}$  \\ 
\\
\multirow{3}{*}{sigmoidal mixed model} & \multirow{3}{*}{$y=\frac{\alpha\beta(\sum_{i\in {\mathrm{pa}}}x_{i} + \lambda\epsilon)}{1 + \lvert \beta (\sum_{i\in {\mathrm{pa}}}x_{i} + \lambda\epsilon + \gamma) \rvert}$} & $\alpha \sim \mathrm{Exponential}(0.25) + 1$ \\
 &  & $ \beta \sim \mathrm{Uniform([-1, -0.25] \cup [0.25, 1])}$ \\ 
 &  & $\gamma \sim \mathrm{Uniform([-2, 2])}$ \\
\\
\multirow{3}{*}{sigmoidal additive model} & \multirow{3}{*}{$y=\sum_{i\in {\mathrm{pa}}}\frac{\alpha\beta(x_{i}+\gamma)}{1+\lvert \beta(x_{i}+ \gamma) \rvert} + \lambda\epsilon$} & $\alpha \sim \mathrm{Exponential}(0.25) + 1$ \\
 &  & $ \beta \sim \mathrm{Uniform([-1, -0.25] \cup [0.25, 1])}$ \\ 
 &  & $\gamma \sim \mathrm{Uniform([-2, 2])}$ \\

\bottomrule

\end{tabular*}
{\footnotesize }
\end{table*}

\subsection{Hyperparameter choices}

For the Lagrangian optimization, we run the optimization for 100 rounds for graphs up to 6 nodes and 200 rounds for graphs with $7, 8$ or $9$ nodes, with 30 optimization steps per round. The learning rate is set to 0.3, and the acyclicity constraint used is the ``DAGMA'' constraint as defined by \citet{bello2022dagma}. The augmented Lagrangian parameters include an initial $\lambda$ of 2, an initial $\tau$ of 0.1, a maximum $\tau$ of 4, and a $\gamma$ of 1.2 (for the notation as defined in \cref{app:lagrangian}). For the non-linear case, a multilayer perceptron (MLP) must be learned at each step to estimate the value of the causal query from the data once we have the adjustment set. This MLP has $1$ hidden layer of dimension $32$, and is trained for $1000$ epochs with a learning rate of $0.05$ and using the Adam optimizer~\citep{kingma2020method}. For the probabilistic DAG method, we use the same number of rounds as for the Lagrangian, with a learning rate of $0.0001$. For both methods, the initial temperature for Gumbel-Softmax sampling is set at $1$, decaying at $0.9997$ per step. 

All of these hyperparameter settings were maintained consistently across nearly 10,000 simulations. While some real-world applications may require tailored adjustments, these largely untuned parameters reliably produced robust results throughout our diverse experimental settings. 

\Cref{tab:grid-random} shows the parameters for our data generation over which we built a Euclidean grid for our experiments, resulting in $5670$ simulations for the case with random uncertainty and $3150$ simulations for PC uncertainty.

\begin{table}[t]
    \centering
    \begin{tabular}{lll}
        \toprule
        \textbf{Parameter} & \textbf{Random uncertainty} & \textbf{PC uncertainty} \\
        \midrule
        number of nodes & [3, 4, 5, 6, 7, 8, 9] & [3, 4, 5, 6, 7, 8, 9] \\
        sure edge probability & [0.3, 0.5, 0.7] & N/A \\
        forbidden edge probability & [0.3, 0.5, 0.7] & N/A \\
        causal mechanism & ['linear', 'sig add', 'sig mix'] & ['linear', 'sig add', 'sig mix'] \\
        random seed & [0, 17, 34, 51, 68] & [0, 17, 34, 51, 68] \\
        edge probability & [0.3, 0.5, 0.7] & [0.3, 0.5, 0.7] \\
        adjustment type & [parent, optimal] & [parent, optimal] \\
        number of permutations & N/A & [3, 5, 10, 15, 20] \\
        \bottomrule
    \end{tabular}
    \caption{Characterization of the different data generation settings used in our experiments. There are a total of $5670$ simulations for the case with random uncertainty and $3150$ simulations for PC uncertainty. Additionally, for simulations for more than 9 nodes, we only vary the sure edge probability, the forbidden edge probability, and the random seed (to 3 values), giving us a total of 27 simulations each for 10 to 25 nodes.}
    \label{tab:grid-random}
\end{table}

\begin{algorithm}
\caption{Linear additive model (Lagrangian)}\label{alg:linear}
\begin{algorithmic}[1]

\State \textbf{Input} sure edges list, forbidden edges list, index of treatment variable $x$, index of effect variable $y$, number of variables $d$, dataset $D$
\State \textbf{Output} maximum and minimum of causal query $Q(\hat{A}_{\alpha})$
\State \textbf{Initialization} initialization of $d \times d$ parameter matrix

\State \textbf{Loop} $k$ iterations
    \State \quad Sample both soft and hard adjacency matrix $\hat{A}_{\mathrm{hard}}$ and $\hat{A}_{\mathrm{soft}}$ using Gumbel-Softmax, assign the edges to $1$ if the edge is in the sure edges list, to $0$ if the edge is in the forbidden edges list
    \State \quad Apply straight-through trick to adjacency matrix $\hat{A}_\alpha \gets (\hat{A}_{\mathrm{hard}}-\hat{A}_{\mathrm{soft}}).\mathrm{detach()} + \hat{A}_{\mathrm{soft}}$
    \State \quad Get adjustment set $Z(\hat{A}_\alpha,x,y)$ of $x$ on $y$ 
    \State \quad Compute $\cQ(\hat{A}_\alpha)$ by regressing $D[y]$ on $D[Z]$ and $D[x]$
    \State \quad Minimize $\cL(\hat{A}_\alpha, \lambda, \tau) \gets \pm \cQ(\hat{A}_\alpha) +  \xi(h(\hat{A}_\alpha), \lambda, \tau)$

\end{algorithmic}
\end{algorithm}

\begin{algorithm}[t]
\caption{Non-linear model (Lagrangian)}\label{alg:nonlinear}
\begin{algorithmic}[1]

\State \textbf{Input} sure edges list, forbidden edges list, index of treatment variable $x$, index of effect variable $y$, number of variables $d$, dataset $D$, number of maximum iteration $i==MAX\_ITER$
\State \textbf{Output} maximum and minimum of causal query $\cQ(\hat{A}_{\alpha})$
\State \textbf{Initialization} initialization of $d \times d$ parameter matrix

\State \textbf{Loop} $k$ iteration
    \State \quad Sample both soft and hard adjacency matrix $\hat{A}_{\mathrm{hard}}$ and $\hat{A}_{\mathrm{soft}}$ using Gumbel-Softmax, assign the edges to $1$ if the edge is in the sure edges list, to $0$ if the edge is in the forbidden edges list
    \State \quad Apply straight-through trick to adjacency matrix $\hat{A}_{\alpha} \gets (\hat{A}_{\mathrm{hard}}-\hat{A}_{\mathrm{soft}}).\mathrm{detach()} + \hat{A}_{\mathrm{soft}}$
    \State \quad Get adjustment set $Z(\hat{A}_{\alpha},x,y)$ of $x$ on $y$ 
    \State \quad Initialize an MLP $f$ with input size of $len(Z)+1$ and output size of $1$
    \State \quad \textbf{For} $i \gets 1$ to $MAX\_ITER$ \textbf{do}
    \State \quad \quad Compute $\cL(D,Z,x,y) \gets (D[y]-f(D[x,Z]))^2$
    \State \quad \quad Minimize $\cL(D,Z,x,y)$
    \State \quad \quad \textbf{if} $\cL(D,Z,x,y)$ converged or $i==MAX\_ITER$
    \State \quad \quad \quad \textbf{break}
    \State \quad Freeze the weights of the MLP $f$
    \State \quad Compute $\cQ(\hat{A}_{\alpha}) \gets \mathrm{mean}(f(D[Z],D[x]=1)-f(D[Z],D[x]=0))$ 
    \State \quad Minimize $\cL(\hat{A}_{\alpha}, \lambda, \tau) \gets \pm \cQ(\hat{A}_{\alpha}) +  \xi(h(\hat{A}_{\alpha}), \lambda, \tau)$

\end{algorithmic}
\end{algorithm}

\subsection{Estimation uncertainty}
\label{app:estimation-uncertainty}
It is possible that the ground truth causal query lies on the boundary among all possible values given known edge information. However, even though we know the true causal graph beforehand, the limited data available introduces considerable uncertainty in our statistical estimates. Therefore, we may falsely consider the ground truth causal query to be out of bounds simply because of finite sample estimation errors. To deal with this, we added an additional step after running the optimization over adjacency matrices. For the final matrices achieving the maximum and minimum values of the causal query, we subsample from the dataset $50$ times and estimate the causal query on each of these datasets. Thereby, we obtain a bootstrapped estimation of the variance of our estimated query. In our evaluation, when checking whether the ground truth is within the estimated bounds, we consider the interval $\cQ \in [\hat{\cQ}_{\min} - \sigma(\hat{\cQ}_{\min}), \hat{\cQ}_{\max} + \sigma(\hat{\cQ}_{\max})]$, i.e., extend the point estimates by one standard deviation of the bootstrapped estimates.

\section{Additional experiments}
\label{app:add-experiments}

\begin{figure}
    \centering
    \subfigure[Running time comparison for increasing search space size (\textit{3780} simulations)]{
        \includegraphics[width=0.47\textwidth]{figs/time_comparison_main_nonlinear.pdf}
        \label{fig:runtime-graphs-nl}
    }
    \subfigure[Running time comparison for graphs with different number of nodes (\textit{3780} simulations)]{
        \includegraphics[width=0.47\textwidth]{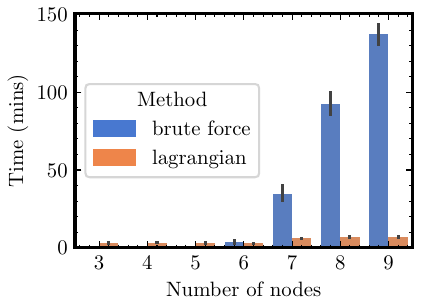} 
        \label{fig:runtime-nodes-nl}
        }
    \caption{Comparison of running times for the Lagrangian and brute force methods for the non-linear mechanisms. }
    \label{app-fig:runtime-nl}
\end{figure}

\begin{figure}
    \centering
    \subfigure[Running time comparison for sure edge probability (\textit{5670} simulations)]{
        \includegraphics[width=0.47\textwidth]{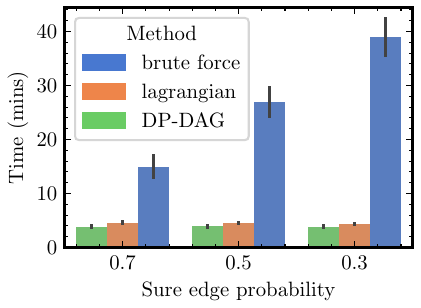}
        \label{fig:runtime-sure-all}
    }
    \subfigure[Running time comparison for forbidden edge probability (\textit{5670} simulations)]{
        \includegraphics[width=0.47\textwidth]{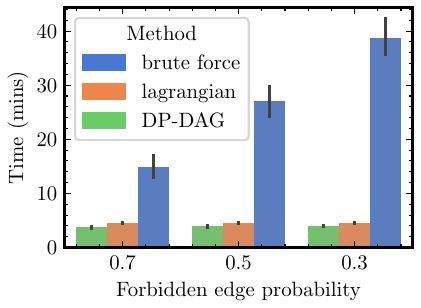} 
        \label{fig:runtime-forbid-all}
        }
    \caption{Comparison of running times for the Lagrangian, DP-DAG and brute force methods for different sure and forbidden edge probabilities used to generate the set of uncertainties. }
    \label{app-fig:runtime-sreforbid-all}
\end{figure}

\Cref{app-fig:runtime-nl} shows a relatively steady running time of the non-linear Lagrangian method as the search space grows, next to a super-exponential increase in the computational complexity of the brute force algorithm. On the right, we see this reflected in the number of nodes.
\cref{app-fig:runtime-sreforbid-all} compares running time for different values of sure and forbidden edge probabilities used to generate the set of uncertainties. The full mechanism for this is described in \cref{app:graph-uncertainty}. As expected, when the sure or forbidden edge probability is low, it implies that a lower proportion of edges from the ground truth are known to be sure or forbidden. This increases the search space and, therefore, the runtime, which is what we see in the plot.

\cref{app-tab:adjustment type} and \cref{app-tab:pc-mechanism} show the metric values for more configurations. In \cref{app-tab:adjustment type}, we see that we have effectively perfect point coverage of the ground truth effect value for both types of adjustment sets. However, we see that DP-DAG is better than the Lagrangian at covering the full bound set. In Bound narrowness, we see that the Lagrangian is better again due to the DP-DAG not being able to account for sure edges in its optimization and, therefore, giving wider bounds as expected.

\begin{table}
    \centering
\begin{tabular}{lrlllrr}
\toprule
 & \multicolumn{2}{c}{\textbf{Point Coverage}} & \multicolumn{2}{c}{\textbf{Bound Coverage}} & \multicolumn{2}{c}{\textbf{Bound Narrowness}} \\ 
           \cmidrule(lr){2-3}\cmidrule(lr){4-5}\cmidrule(lr){6-7}
      \texttt{Adjustment}     & Lagrangian    & DP-DAG   & Lagrangian   & DP-DAG  & Lagrangian   & DP-DAG       \\
\midrule
\texttt{parent} & $0.98 \pm 0.0$ & $\mathbf{1 \pm 0.0}$ & $0.90 \pm 0.0$ & $\mathbf{0.99 \pm 0.0}$ & $\mathbf{2.33 \pm 0.51}$ & $2.69 \pm 0.20$ \\
\texttt{optimal} & $0.98 \pm 0.0$ & $\mathbf{1 \pm 0.0}$ & $0.92 \pm 0.0$ & $\mathbf{0.99 \pm 0.0}$ & $\mathbf{1.99 \pm 0.05}$ & $2.85 \pm 0.22$ \\
\bottomrule
\end{tabular}
\caption{Metric values across different adjustment types for the random uncertainty generation.}
    \label{app-tab:adjustment type}
\end{table}

\begin{table}
    \centering
    \begin{tabular}{lllllrr}
\toprule
   & \multicolumn{2}{c}{\textbf{Point Coverage}} & \multicolumn{2}{c}{\textbf{Bound Coverage}} & \multicolumn{2}{c}{\textbf{Bound Narrowness}} \\ 
           \cmidrule(lr){2-3}\cmidrule(lr){4-5}\cmidrule(lr){6-7}
   \texttt{Mechanism}        & Lagrangian    & DP-DAG   & Lagrangian   & DP-DAG  & Lagrangian   & DP-DAG       \\ \midrule \texttt{linear} & $0.64 \pm 0.01$ & $\mathbf{0.96 \pm 0.01}$ & $0.63 \pm 0.01$ & $\mathbf{0.76 \pm 0.01}$ & $\mathbf{1.25 \pm 0.08}$ & $1.67 \pm 0.09$ \\
\texttt{non-linear} & $0.97 \pm 0.0$ & N/A & $0.90 \pm 0.0$ & N/A & $2.19 \pm 0.05$ & N/A \\
\bottomrule
\end{tabular}
    \caption{Metric values for PC-uncertainty. We see that the values for point coverage are lower because it is not guaranteed in this uncertainty generation that the true bounds actually contain the true value, as described in \cref{app:graph-uncertainty}.}
    \label{app-tab:pc-mechanism}
\end{table}

\section{Implementation details}
\label{app:imp-details}

\subsection{Adjustment sets}
\label{app:adjustment-sets}
Without properly accounting for confounders, any observed association between the treatment and the outcome may reflect not just the treatment's effect but also the confounders' influence. As a special case, parent adjustment takes the parent set of treatment variable $X$; it naturally satisfies the definition of an adjustment set, which blocks all potential back-door paths and, therefore, all non-causal paths. Although parent adjustment has the advantage of easy applicability, it has a larger asymptotic variance, which may lead to imprecise causal effect estimation. This becomes even more inefficient when the parent of treatment shows strong co-linearity with the treatment variable. This inefficiency arises because the parent variables may capture too much variability from treatment, which is unnecessary for estimating the total effect on the outcome. Despite the complexity, an optimal adjustment set (\cref{eq:optimal-set}) includes nodes that are not directed cause of treatment $X$ but add precision variable explaining additional variance in the outcome $Y$, therefore leading to minimum asymptotic estimation variance \citep{Henckel_2022}. In \cref{eq:optimal-set}, $cn$ stands for causal nodes, which are nodes on the directed path from $X$ to $Y$, and forbidden nodes $forb$ are defined as the descendent of the causal path and $X$ itself. 

\begin{equation}
    \label{eq:optimal-set}
    O(X,Y,\cG) = pa(cn(X, Y, \cG))\backslash forb(X,Y,\cG)
\end{equation}

Using an adjustment set to compute causal effect includes a selection procedure that is not differentiable. Therefore, we use a matrix multiplication trick that allows the gradient to backpropagate through the desired entries of the adjacency matrix. We summarize our method in \cref{alg:gradient}.

\begin{algorithm}
\caption{Gradient Preserving Variable Selection}
\label{alg:gradient}
\begin{algorithmic}[1]
\Require Observational dataset \texttt{data}, adjustment set indices \texttt{s} (1D tensor)

\State Create a selection mask for the adjustment variables: \texttt{selection\_mask} $\gets$ \texttt{s}
\State Initialize an empty list for \texttt{mask\_list}

\State \textbf{For }{each index \texttt{idx} in selection\textunderscore mask \textnormal{(non-zero elements)}}
    \State \quad Create a zero matrix \texttt{mult} of size (\texttt{s.shape[0]}, \texttt{s.shape[0]})
    \State \quad Set \texttt{mult[idx[1], idx[1]]} $\gets$ 1
    \State \quad Append \texttt{selection\_mask} $\times$ \texttt{mult} to \texttt{mask\_list}

\State Convert the \texttt{mask\_list} to a 2D tensor along axis 0
\State Compute \texttt{data\_adjustment} $\gets$ \texttt{data} $\times$ \texttt{mask\_list}$^T$ \Comment{Matrix multiplication to select parent columns}

\end{algorithmic}
\end{algorithm}

When using parent adjustment, there is no ambiguity about which edges are blocked. However, this is not the case for the optimal adjustment. When a variable $\Z$ is in the adjustment set, we don't know which edges we need to block (i.e., which entries in the adjacency matrix). $\Z$ may have multiple children on the causal path from $X$ to $Y$. Since estimating the causal effect using an adjustment set does not depend on their children, i.e., edges from the adjustment variable to its children on the causal path are equally good for estimation, we simply take equal contributions from these edges. 

\subsection{Straight-Through Gumble-Softmax distribution}
\label{app:gumble}
In our case, the probability of an edge not existing is defined via
\begin{equation}\label{eq:gumbel-non-exist}
  a_0 = \frac{\exp{((\log(1 - \pi_1)+g_0)/\tau)}}{\exp{((\log(1 - \pi_1)+g_0)/\tau)} + \exp{((\log(\pi_1)+g_1)/\tau)}}\:,
\end{equation}
where $\pi_1$ is the probability of an edge being $1$, $g_0$ and $g_1$ are drawn from $\mathrm{Gumbel}(0,1)$ distribution~\citep{gumbel1935valeurs, gumbel1941return}, and $\tau$ is the temperature parameter. When $\tau$ approaches $0$, the distribution becomes one-hot. By contrast, when $\tau$ becomes large, the distribution tends to be uniform. During optimization, we take a large initial value for $\tau$ for better searching and anneal the temperature in order to obtain a static adjacency matrix. 
Similarly, if an edge exists, the probability can be written as:
\begin{equation}\label{eq:gumbel-exist}
  a_1 = \frac{\exp{((\log(\pi_1)+g_1)/\tau)}}{\exp{((\log(1 - \pi_1)+g_0)/\tau)} + \exp{((\log(\pi_1)+g_1)/\tau)}}
\end{equation}

\subsection{Lagrangian optimization}
\label{app:lagrangian}
With the notation as defined in the paper, the Lagrangian we aim to minimize with respect to $\alpha$ can be formulated as:
\begin{equation}\label{eq:subproblem}
  \cL(\alpha, \lambda, \tau) := \pm \cQ(\alpha) +  \xi(h(\alpha), \lambda, \tau) \;\;\;\;\;\text{ with }\;\;\; \xi(h(\alpha), \lambda, \tau) := - \lambda h(\alpha) + \frac{\tau h(\alpha)^2}{2}
\end{equation}
where $-$/$+$ is used for the upper/lower bound. $\tau$ is increased throughout the optimization procedure and is seen as a temperature parameter.

\begin{table}[h]
\centering
\caption{Overview of resources used in our work.}\label{tab:software}
\smallskip
\begin{tabular*}{\linewidth}{@{\extracolsep{\fill}} l l l}
  \toprule
    \textbf{Name} & \textbf{Reference} & \textbf{License}  \\ 
    \midrule
    Python          &  \cite{vanrossum2009python}
                    & \href{https://docs.python.org/3/license.html#psf-license}{PSF License} \\
    PyTorch         &  \cite{paszke2019pytorch}
                    & \href{https://github.com/pytorch/pytorch/blob/main/LICENSE}{BSD-style license} \\
    Numpy           &  \cite{harris2020numpy}
                    & \href{https://github.com/numpy/numpy/blob/main/LICENSE.txt}{BSD-style license} \\
    Pandas          &  \cite{reback2020pandas,mckinney-proc-scipy-2010}
                    & \href{https://github.com/pandas-dev/pandas/blob/main/LICENSE}{BSD-style license} \\
    Matplotlib      &  \cite{hunter2007matplotlib}
                    & \href{https://matplotlib.org/stable/users/project/license.html}{modified PSF} \\
    Scikit-learn    &  \cite{pedregosa2011scikit}
                    & \href{https://github.com/scikit-learn/scikit-learn/blob/main/COPYING}{BSD 3-Clause} \\
    SciPy           &  \cite{virtanen2020scipy}
                    & \href{https://github.com/scipy/scipy/blob/main/LICENSE.txt}{BSD 3-Clause} \\
    SLURM           & \cite{yoo2003slurm}
                    & \href{https://github.com/SchedMD/slurm/tree/master?tab=License-1-ov-file}{modified GNU GPL v2} \\
    networkx        & \cite{networkx}
                    & \href{https://raw.githubusercontent.com/networkx/networkx/master/LICENSE.txt}{BSD 3-Clause} \\
    DCDI            & \cite{Brouillard2020differentiable} 
                    & \href{https://github.com/slachapelle/dcdi/blob/master/LICENSE.md}{MIT license} \\
    optimaladj      & \cite{Smucler2021, smucler2022noteefficientminimumcost}
                    &\href{https://github.com/facusapienza21/optimaladj?tab=readme-ov-file} {MIT license} \\
    \bottomrule
\end{tabular*}
\end{table}

Given an approximate minimum $\alpha$ of this subproblem, we then update $\lambda$ and $\tau$ according to $\lambda \gets \max\{0, \lambda - \tau h(\alpha)\}$ and $\tau \gets \alpha \cdot \tau$ for a fixed $\alpha > 1$.
The overall strategy is to iterate between minimizing \cref{eq:subproblem} and updating $\lambda_l$ and $\tau$. Separately solving this optimization, once for a minimization problem and once for a maximization problem, would give us the bounds. 

\textbf{Enforcing sure and forbidden edges}: For the Lagrangian method, we enforce sure and forbidden edges by blocking the gradients through masking. We construct a masking matrix $M$ of size $d \times d$ with $M_{ij} = 0$ if $(i, j) \in \bE_s \cup \bE_f$ and $M_{ij} = 1$ otherwise. Each time a graph is sampled, we mask out the sure and forbidden edges in the adjacency matrix with the mask and replace the sure edges with $1$. For the DP-DAG method, only forbidden edges are ensured, therefore, the masking matrix $M$ is defined with $M_{ij} = 0$ if $(i,j) \in \bE_f$. Similarly, after a DAG is sampled, we apply the mask to its adjacency matrix to remove the forbidden edges.

\section{Resources}
\label{app:code}

Our project heavily relies on available open-source software packages and data sources, which we list in \cref{tab:software}.
\end{document}